\documentclass[10pt,twocolumn,letterpaper]{article}

\usepackage{wacv}              % To produce the CAMERA-READY version
\usepackage[accsupp]{axessibility}
\usepackage{graphicx}
\usepackage{amsmath}
\usepackage{amssymb}
\usepackage{booktabs}

\usepackage[breaklinks,colorlinks]{hyperref}

\usepackage{multirow}
\usepackage{multicol}
\usepackage{pifont}
\usepackage{dsfont}
\usepackage{xcolor}
\newcommand{\blue}[1]{\textcolor{blue}{#1}}

\usepackage[capitalize]{cleveref}
\crefname{section}{Sec.}{Secs.}
\Crefname{section}{Section}{Sections}
\Crefname{table}{Table}{Tables}
\crefname{table}{Tab.}{Tabs.}

\def\wacvPaperID{2579} % *** Enter the WACV Paper ID here
\def\confName{WACV}
\def\confYear{2025}

\makeatletter
\newcommand{\printfnsymbol}[1]{%
  \textsuperscript{\@fnsymbol{#1}}%
}
\makeatother

\title{\texttt{CamoFA}: A Learnable Fourier-based Augmentation for\\ Camouflage Segmentation}

\author {
    Minh-Quan Le\textsuperscript{\rm 1, 2, 3}\thanks{Equal contribution} ,
    Minh-Triet Tran\textsuperscript{\rm 1, 2, 4}\printfnsymbol{1},
    Trung-Nghia Le\textsuperscript{\rm 1, 2},
    Tam V. Nguyen\textsuperscript{\rm 5},
    Thanh-Toan Do\textsuperscript{\rm 6}\\
    \\
    \textsuperscript{\rm 1} University of Science, Ho Chi Minh City, Vietnam\\
    \textsuperscript{\rm 2} Vietnam National University, Ho Chi Minh City, Vietnam \\
    \textsuperscript{\rm 3} Stony Brook University, United States\\    
    \textsuperscript{\rm 4} John von Neumann Institute, Ho Chi Minh City, Vietnam \\
    \textsuperscript{\rm 5} University of Dayton, United States\\
    \textsuperscript{\rm 6} Monash University, Australia 
}
\begin{document}

\maketitle

%%%%%%%%% ABSTRACT
\begin{abstract}
Camouflaged object detection (COD) and camouflaged instance segmentation (CIS) aim to recognize and segment objects that are blended into their surroundings, respectively. While several deep neural network models have been proposed to tackle those tasks, augmentation methods for COD and CIS have not been thoroughly explored. Augmentation strategies can help improve models' performance by increasing the size and diversity of the training data and exposing the model to a wider range of variations in the data. Besides, we aim to automatically learn transformations that help to reveal the underlying structure of camouflaged objects and allow the model to learn to better identify and segment camouflaged objects. To achieve this, we propose a learnable augmentation method in the frequency domain for COD and CIS via the Fourier transform approach, dubbed \texttt{CamoFA}. Our method leverages a conditional generative adversarial network and cross-attention mechanism to generate a reference image and an adaptive hybrid swapping with parameters to mix the low-frequency component of the reference image and the high-frequency component of the input image. This approach aims to make camouflaged objects more visible for detection and segmentation models. Without bells and whistles, our proposed augmentation method boosts the performance of camouflaged object detectors and instance segmenters by large margins.
\end{abstract}

%%%%%%%%% BODY TEXT
\section{Introduction}
% \label{sec:intro}
\label{sec1}

Camouflage refers to using any combination of materials, coloration, or illumination for concealment, either by making animals or objects hard to see or by disguising them as something else. In the context of computer vision, camouflaged object detection (COD) and camouflaged instance segmentation (CIS) are tasks that aim to recognize and segment objects that are integrated into their surroundings. COD is the task of identifying objects that are ``seamlessly'' embedded in their surroundings \cite{LE-CVIU2019}. The high intrinsic similarities between the target object and the background make COD far more challenging than traditional object detection tasks \cite{Fan-CVPR2020}. CIS is a related task that involves not only detecting camouflaged objects but also segmenting them at the pixel level \cite{ltnghia-TIP2022}. Both COD and CIS require sophisticated computer vision algorithms that can distinguish between the target object and its background despite their high visual similarity. These tasks have several potential applications, including wildlife monitoring, search, and rescue operations \cite{LE-CVIU2019},  medical diagnosis such as polyp segmentation \cite{Fan-MICCAI2020}, COVID-19 identification from lung x-rays \cite{Pham_2024_WACV, pham2024fgcxrradiologistalignedgazedataset}.

\begin{figure}[!t]
    \centering
    \includegraphics[width=\linewidth]{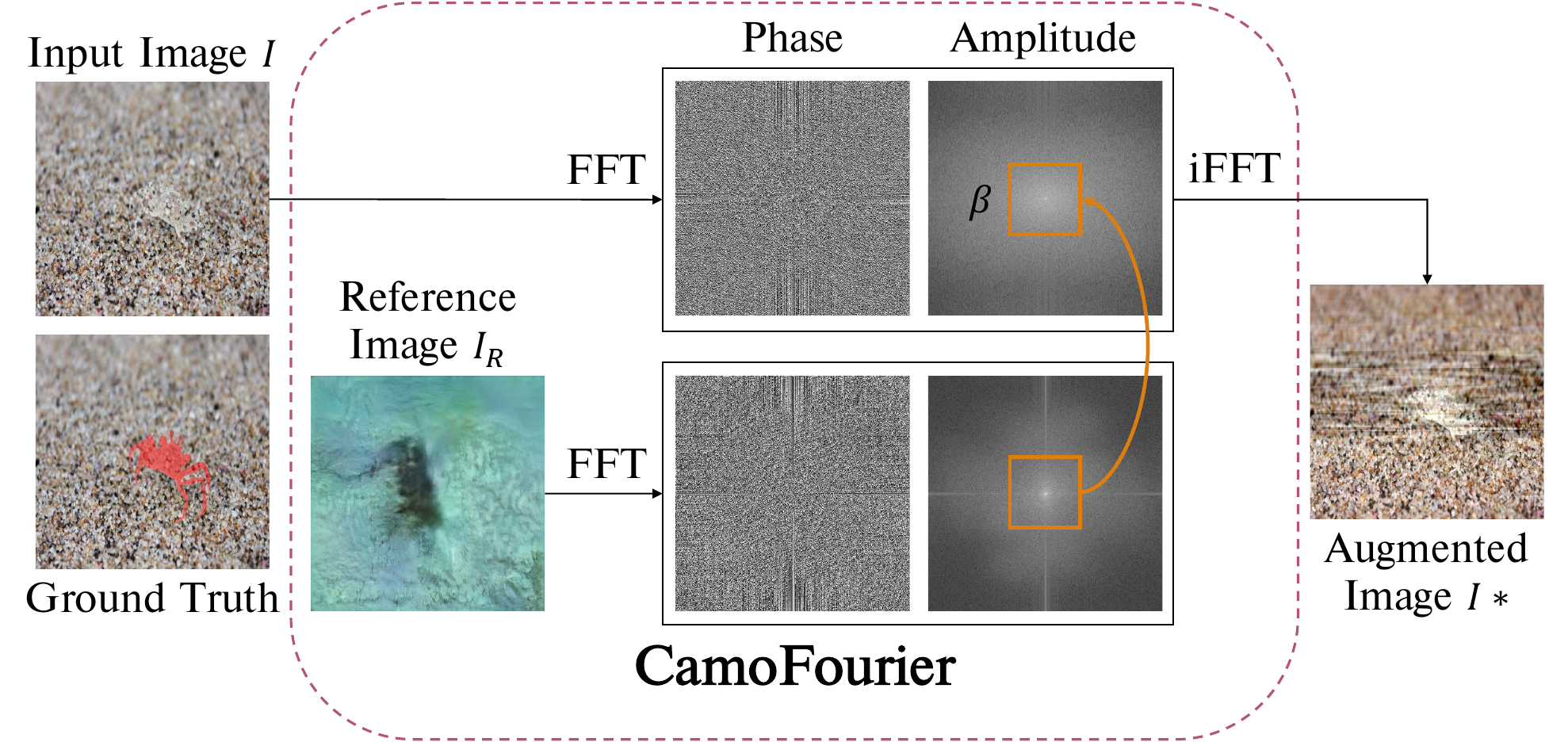}
    \caption{
    Our \texttt{CamoFA} not only preserves the spatial structure of an image but also highlights the underlying structure of camouflaged objects for better identification and segmentation.}
    \label{fig:augmentation}
    \vspace{-2mm}
\end{figure}

The performance of COD has recently been elevated by convolutional neural networks (CNNs)-based approaches. Current cutting-edge methods concentrate on designing detector architectures based on attention \cite{Pang-CVPR2022, Fan-CVPR2020, Jia-CVPR2022}, and specific features of camouflaged objects \cite{Fan-TPAMI2022, Zhai-CVPR2021}. Moreover, OSFormer \cite{Pei-ECCV2022} leverages ViT \cite{Dosovitskiy-ICLR2021} to propose the first architecture for the CIS task. 
While several architectures have been designed to address COD and CIS tasks, augmentation strategies have not been completely investigated, even though they can help improve the model’s ability to generalize to new data and increase its robustness to variations in the input. Therefore, this work explores the data-centric approach to COD and CIS problems by proposing a learnable augmentation method in the frequency domain, named \texttt{CamoFA}, which allows deep learning models to learn how to augment their own training data. In this way, our learnable augmentation aims to reveal the underlying structure of camouflaged objects and allow the model to learn to better identify and segment camouflaged objects. 

Although several augmentation methods have been proposed for generic object detection \cite{Zhang-ICLR2018, Yun-ICCV2019, Zoph-ECCV2020} and instance segmentation \cite{Ghiasi-CVPR2021} tasks, these works could not be directly applied to COD and CIS tasks as these methods introduce occlusions and deformations to input images, which may make camouflaged objects even harder to detect and segment \cite{Zhang-ICLR2018}. Moreover, these methods do not consider the intrinsic similarities between the target object and the background \cite{Ghiasi-CVPR2021}, the main challenges for COD and CIS. 
Therefore, we introduce a learnable augmentation strategy in the frequency domain via Fourier transform for COD and CIS. The Fourier transform of an image can encode semantic information and appearance information of the image in its phase and amplitude components, respectively. Changing the amplitude information while keeping the phase information of the Fourier transform of an image can change the appearance of the image, but it is still recognizable. Additionally, Fourier transform can also preserve the spatial structure and resolution of an image, see Figure~\ref{fig:augmentation}. 

Motivated by the above properties of the Fourier transform, we propose a learnable augmentation approach that aims to manipulate the amplitude information of the Fourier transform of an input image to enhance the visibility of camouflaged objects in the image. 
In particular, our method leverages a conditional generative adversarial network to synthesize a generated image $\mathcal{I}_G$ and cross-attention mechanism to learn the spatial correspondence and alignment between the input image $\mathcal{I}$ and the generated one $\mathcal{I}_G$, which outputs a reference image $\mathcal{I}_R$. Furthermore, we propose a hybrid swapping with an adaptive parameter to mix the low-frequency component of the reference image and the high-frequency component of the input image to control the amount of texture and color information that is transferred from the reference image $\mathcal{I}_R$ to the input image $\mathcal{I}$. To the best of our knowledge, our work is the first one that designs a trainable Fourier-based augmentation specifically for both COD and CIS tasks. 
In summary, our main contributions are three-fold.
\begin{itemize}
    \item We propose a novel learnable Fourier-based augmentation for camouflaged object detection and instance segmentation, highlighting the camouflaged object of interest from the background and making it more visible for deep models. The augmentation framework is flexible and can be adapted to different segmentation or detection algorithms. 
    \item We introduce an adaptive hybrid swapping approach that can control the amount of texture and color information transferred from the reference image to the input image. We leverage a simple yet effective cross-attention mechanism between the input and generated image so that the reference image can transfer its texture and color information to the input image in a more attentive manner.
    \item Extensive experimental results on various camouflaged object detection and instance segmentation methods on different datasets show that the proposed method significantly improves performance of existing models.
\end{itemize}

\section{Related Work}

\textbf{Camouflaged object detection.} Fan \etal \cite{Fan-CVPR2020} present Search and Identification Net (SINet), a strong baseline for the COD task. There is also an enhanced version of SINet called SINetV2 \cite{Fan-TPAMI2022} with two well-elaborated sub-modules: neighbor connection decoder (NCD) and group-reversal attention (GRA). To well involve the frequency clues in the CNN models, Zhong \etal \cite{Zhong-CVPR2022} introduce the frequency domain as an additional clue to detect camouflaged objects from backgrounds better. Several sophisticated architectures have been proposed for COD, our \texttt{CamoFA} is a learnable augmentation strategy for COD that can be plugged into these existing works and improve their performance in an efficient and easy implementation.
\begin{figure*}[!t]
    \centering
    \includegraphics[width=\textwidth]{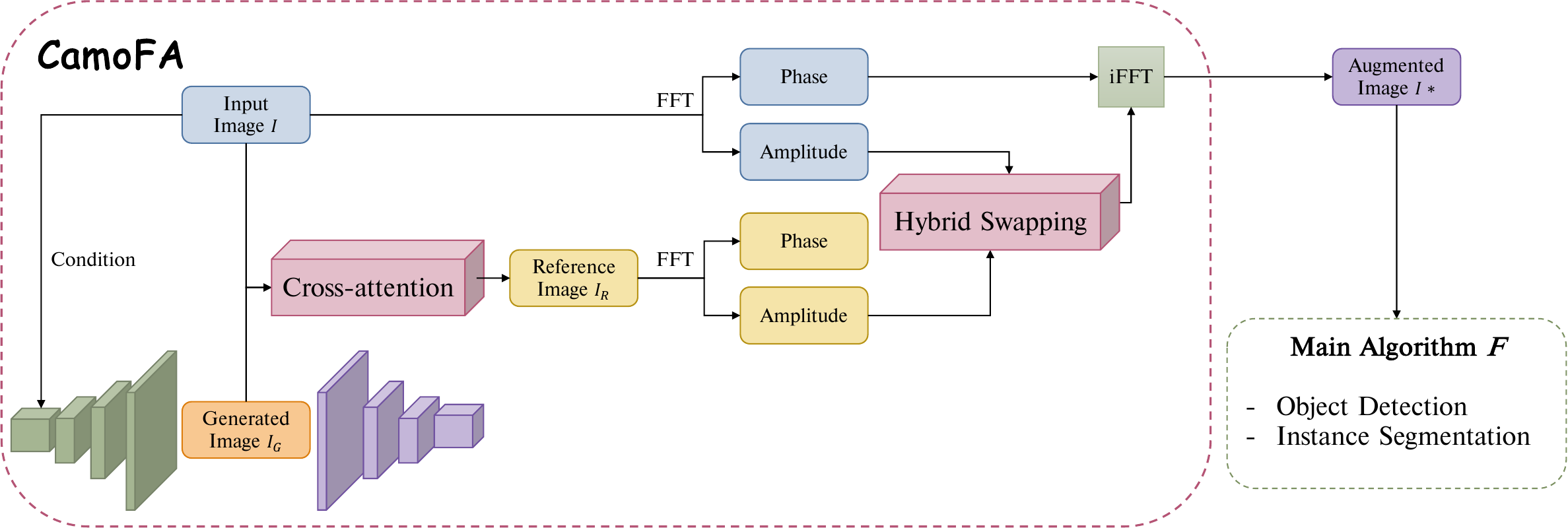}
    \caption{Overview of the proposed \texttt{CamoFA}. Our method leverages a conditional generative adversarial network and cross-attention mechanism to generate a reference image and an adaptive hybrid swapping with parameters to mix the low-frequency component of the reference image and the high-frequency component of the input image.}
    \label{fig:framework}
    \vspace{-2mm}
\end{figure*}

\textbf{Camouflaged instance segmentation.} Le \etal \cite{ltnghia-TIP2022} investigate the interesting yet challenging problem of camouflaged instance segmentation. To promote the new task of camouflaged instance segmentation of in-the-wild images, Le \etal \cite{ltnghia-TIP2022} introduce a dataset, dubbed CAMO++, that extends the preliminary CAMO dataset (camouflaged object segmentation) in terms of quantity and diversity. To detect and segment the whole scope of a camouflaged object, camouflaged object detection is introduced as a binary segmentation task, with the binary ground truth camouflage map indicating the exact regions of the camouflaged objects. Pei \etal~present OSFormer \cite{Pei-ECCV2022}, the first one-stage transformer framework for camouflaged instance segmentation. As the CIS task has not been deeply explored, our \texttt{CamoFA} is an efficient plug-and-play tool that helps improve CIS models' performance and is valuable for researchers working on CIS.

\textbf{Data augmentation.} Data augmentation is a critical component of training deep learning models for object detection and segmentation tasks \cite{Zhang-ICLR2018, Yun-ICCV2019, Zoph-ECCV2020, Ghiasi-CVPR2021, Luo-CamDiff2023, le2024maskdiff, le2023gunnelguidedmixupaugmentation, nguyen2024instanceawaregeneralizedreferringexpression, nguyentruong2024visionawaretextfeaturesreferring}. Data augmentation has significantly improved generic object detection and instance segmentation, but its potential has not been thoroughly investigated for camouflaged object detection and camouflaged instance segmentation. The existing works are augmentation methods in the spatial domain. When images are augmented in the spatial domain, it is possible to introduce occlusions, deformations, or noise to the input images, which may make the camouflaged objects even harder to detect and segment.  Moreover, these methods do not consider the intrinsic similarities between the target object and the background, which are the main challenges for camouflaged object detection and camouflaged instance segmentation. Our \texttt{CamoFA} is an augmentation method in the frequency domain specifically designed for COD and CIS and can help avoid these artifacts by preserving spatial information.

It is worth noting that Fourier transform has been leveraged for domain generalization \cite{Xu-CVPR2021}, and domain adaptation \cite{Yang-CVPR2020}. Unlike those works, in this paper, we explore Fourier transform to propose a learnable augmentation method for COD and CIS tasks.

\section{Proposed Method}

\subsection{Overview Framework}
We propose a general framework (Figure~\ref{fig:framework}) that can be used to provide an effective and highly accurate solution for any segmentation or detection algorithm for camouflage objects. Specifically, we propose a learnable data augmentation module that can be integrated to be \textit{end-to-end training} with any existing object detection and segmentation methods. In this framework, the original input image $\mathcal{I}$ is transformed to synthesize a new input image $\mathcal{I*}$ that can highlight the camouflaged object of interest from the background and is more suitable for the main processing algorithm $\mathcal{F}$, including segmentation and detection. 
It is worth noting that the proposed module is trained together with the main algorithm $\mathcal{F}$ in an end-to-end fashion. 

As shown in Figure~\ref{fig:framework}, we aim to synthesize the transformed image  $\mathcal{I}*$ from an original input image $\mathcal{I}$ by replacing the amplitude in the FFT of $\mathcal{I}$ with the amplitude of the FFT of a context-aware reference image $\mathcal{I}_R$. The input image $\mathcal{I}$ is used as the conditional information for a generative model to generate an RGB image $\mathcal{I}_G$. Then, both the generated image  $\mathcal{I}_G$ and original input image  $\mathcal{I}$ are fed into a Cross Attention module to produce a context-aware reference image  $\mathcal{I}_R$ capturing information of both $\mathcal{I}$ and $\mathcal{I}_G$.  

We intend to synthesize $\mathcal{I}_R$ so that its amplitude component is more appropriate than that of the original image  $\mathcal{I}$ to differentiate the camouflaged object and its surrounding area. Then, we blend the original image $\mathcal{I}$ with the context-aware reference image  $\mathcal{I}_R$ by swapping the amplitude component from the FFT of $\mathcal{I}_R$ with the one of the original image  $\mathcal{I}$. 
We keep the phase component from the FFT of $I$ because we want to preserve the spatial structure and the semantic information of the original input image. We are inspired by the idea of swapping the amplitude component for style transfer \cite{Yang-CVPR2020}, and adopt this idea into our proposed framework. However, in our solution, instead of using a target style for reference as in \cite{Yang-CVPR2020}, we train our framework with data to generate a reference image $\mathcal{I}_R$ that provides a better FFT amplitude component for camouflaged object detection or segmentation.

We do not simply swap entirely the amplitude from a reference image $\mathcal{I}_R$ to the original image $\mathcal{I}$, but we propose an adaptive hybrid swapping approach. 
We reuse the amplitude at high frequencies in the FFT of the original image $\mathcal{I}$ because the high-frequency components are responsible for the texture and detail of an image and adopt the amplitude at low frequencies in the FFT of the reference image $\mathcal{I}_R$ for transferring the texture and color information from the reference image. 

Furthermore, we do not set a fixed threshold frequency for the hybrid swapping step but we decide to train an adaptive threshold for hybrid swapping. In this way, the amplitude swapping step can exploit adaptively to the original input image $\mathcal{I}$.

\subsection{Learnable Fourier-based Augmentation}
\textbf{GAN module.}
Let $G$ be the generator network and $D$ be the discriminator network. The generator $G$ takes an input image $\mathcal{I}$ and a random noise vector $\mathbf{z}$ as inputs and outputs $\hat{\mathcal{I}_G}=G(\mathcal{I},\mathbf{z})$, which is an image that is supposed to look realistic given $\mathcal{I}$. The discriminator $D$ takes the concatenation of $(\mathcal{I},\mathcal{I}_G)$ as input and outputs $D(\mathcal{I},\mathcal{I}_G)$, which is a scalar value indicating how likely $\mathcal{I}_G$ is to be a realistic image given $\mathcal{I}$. The objective function of the conditional generative adversarial network (cGAN) is given by:
\begin{align*}
&\mathcal{L}_{cGAN}(G, D) = \mathbb{E}_{\mathcal{I} \sim p_{\text {data }}(\mathcal{I})}[\log D(\mathcal{I}, \mathcal{I})] \\
&+\mathbb{E}_{\mathcal{I} \sim p_{\text {data }}(\mathcal{I}), \mathbf{z} \sim p_{\mathbf{z}}(\mathbf{z})}\left[\log (1-D(\mathcal{I}, G(\mathcal{I}, \mathbf{z})))\right].
\end{align*}
 
Similar to the cycle-consistency loss used in some image-to-image translation models \cite{Isola-CVPR2017}, we use a conditional generative adversarial network with an additional $L1$ loss term to measure the similarity between the generated image $\mathcal{I}_G$ and the input image $\mathcal{I}$.  $L1$ loss can be written as: 
\begin{equation}
\mathcal{L}_{L1}(G) = \mathbb{E}_{\mathcal{I} \sim p_{\text {data }}(\mathcal{I}), \mathbf{z} \sim p_{\mathbf{z}}(\mathbf{z})}\left[\|\mathcal{I}-G(\mathcal{I},\mathbf{z})\|_1\right],
\end{equation}
where $\|\cdot\|_1$ denotes the $L1$ norm. The total objective function of your model can be written as:
\begin{equation}
\min _G \max _D \mathcal{L}_{cGAN}(G, D) + \lambda \mathcal{L}_{L1}(G),
\end{equation}
where $\lambda$ is a hyperparameter that controls the weight of the $L1$ loss. By adding this term, we encourage the generator to produce images that are not only realistic but also close to the input in %pixel space. 
the spatial domain (RGB domain). 

\textbf{Basic swapping of amplitude.\label{paragraph:basic_swap}} After transforming both input image $\mathcal{I}$ and reference image $\mathcal{I}_R$ (produced by the Cross Attention module) to the frequency domain via FFT, we obtain the phase and amplitude of each image. Then we swap the amplitude of both images. The phase of the input image together with the amplitude of the reference image is transformed back to the spatial domain via inverse FFT (iFFT) to get $\mathcal{I}*$. By swapping the amplitude of the input and reference images, we transfer the texture and color information from the reference image to the input image, while preserving the shape and structure information from the input image. 

Let $\mathcal{F}(\mathcal{I})$ and $\mathcal{F}(\mathcal{I}_R)$ be the Fourier transforms of $\mathcal{I}$ and $\mathcal{I}_R$, respectively. We define $\mathcal{A}(\mathcal{F}(\mathcal{I}))$ and $\phi(\mathcal{F}(\mathcal{I}))$ to be the  amplitude and phase of $\mathcal{F}(\mathcal{I})$. Similarly, the amplitude and phase of $\mathcal{F}(\mathcal{I}_R)$ are denoted by $\mathcal{A}(\mathcal{F}(\mathcal{I}_R))$ and $\phi(\mathcal{F}(\mathcal{I}_R))$. 

The inverse Fourier transform of the amplitude-swapped image is given by:
\begin{equation}
\mathcal{I}* = \mathcal{F}^{-1}(\mathcal{A}(\mathcal{F}(\mathcal{I}_R)), \phi(\mathcal{F}(\mathcal{I}))).
\end{equation}

The image $\mathcal{I}*$ is the output of the amplitude swapping operation. It is the input image $\mathcal{I}$ with the texture and color information from the reference image $\mathcal{I}_R$.

\begin{table*}[!t]
\centering
\footnotesize
\resizebox{1\textwidth}{!}{
\begin{tabular}{lcccccccccccc}
\toprule
\multirow{2}{*}{\textbf{Method}} & \multicolumn{4}{c}{\textbf{NC4K-Test}} & \multicolumn{4}{c}{\textbf{COD10K-Test}} &  \multicolumn{4}{c}{\textbf{CAMO-Test}} \\
 &
  S$_m\uparrow$ &
  $\alpha$E $ \uparrow$ &
  $w$F $ \uparrow$ &
  M $ \downarrow$ & 
  S$_m \uparrow$ &
  $\alpha$E $ \uparrow$ &
  $w$F $ \uparrow$ &
  M $ \downarrow$ &
  S$_m \uparrow$ &
  $\alpha$E $ \uparrow$ &
  $w$F $ \uparrow$ &
  M $ \downarrow$ \\
\midrule
SINet \cite{Fan-CVPR2020} & 0.808 & 0.883 & 0.723 & 0.058 & 0.776 & 0.867 & 0.631 & 0.043 & 0.745 & 0.825 & 0.644 & 0.092 \\
SINet \cite{Fan-CVPR2020} + \textbf{\texttt{CamoFA}} & \blue{0.820} & \blue{0.894} & \blue{0.746} & \blue{0.055} & \blue{0.782} & \blue{0.881} & \blue{0.652} & \blue{0.040} & \blue{0.773} & \blue{0.861} & \blue{0.665} & \blue{0.087} \\
SINetV2 \cite{Fan-TPAMI2022} & 0.847 & 0.898 & 0.770 & 0.048 & 0.815 & 0.863 & 0.680 & 0.037 & 0.820 & 0.875 & 0.743 & 0.070 \\ 
SINetV2 \cite{Fan-TPAMI2022} + \textbf{\texttt{CamoFA}} & \blue{0.859} & \blue{0.914} & \blue{0.778} & \blue{0.043} & \blue{0.826} & \blue{0.890} & \blue{0.704} & \blue{0.032} & \blue{0.858} & \blue{0.893} & \blue{0.752} & \blue{0.066} \\ 
SegMaR \cite{Jia-CVPR2022} & 0.841 & 0.905 & 0.781 & 0.046 & 0.833 & 0.895 & 0.724 & 0.033 & 0.815 & 0.872 & 0.742 & 0.071 \\
SegMaR \cite{Jia-CVPR2022} + \textbf{\texttt{CamoFA}} & \blue{0.859} & \blue{0.927} & \blue{0.803} & \blue{0.045} & \blue{0.862} & \blue{0.916} & \blue{0.744} & \blue{0.029} & \blue{0.828} & \blue{0.894} & \blue{0.771} & \blue{0.068} \\
ZoomNet \cite{Pang-CVPR2022} & 0.853 & 0.907 & 0.784 & 0.043 & 0.838 & 0.893 & 0.729 & 0.029 & 0.820 & 0.883 & 0.752 & 0.066 \\
ZoomNet \cite{Pang-CVPR2022} + \textbf{\texttt{CamoFA}} & \blue{0.872} & \blue{0.923} & \blue{0.801} & \blue{0.037} & \blue{0.864} & \blue{0.911} & \blue{0.740} & \blue{0.025} & \blue{0.852} & \blue{0.906} & \blue{0.774} & \blue{0.060} \\
FDNet \cite{Zhong-CVPR2022} & 0.834 & 0.895 & 0.750 & 0.052 & 0.837 & 0.897 & 0.731 & 0.030 & 0.844 & 0.903 & 0.778 & 0.062 \\ 
FDNet \cite{Zhong-CVPR2022} + \textbf{\texttt{CamoFA}} & \blue{0.857} & \blue{0.926} & \blue{0.794} & \blue{0.048} & \blue{0.840} & \blue{0.918} & \blue{0.752} & \blue{0.029} & \blue{0.863} & \blue{0.927} & \blue{0.790} & \blue{0.055}\\ 
\bottomrule
\end{tabular}
}
\caption{The effectiveness of our \texttt{CamoFA} in improving camouflaged object detection models on three benchmarks: CAMO \cite{LE-CVIU2019}, COD10K \cite{Fan-CVPR2020}, and NC4K \cite{Lv-CVPR2021}. Our \texttt{CamoFA} drastically increases the performance of cutting-edge methods of COD, improved results are highlighted in \blue{blue} color.}
\label{tab:cod_exp}
\vspace{-2mm}
\end{table*}

\subsection{Cross-Attention Module} 

We hypothesize that the reference image $\mathcal{I}_R$ must capture information from the input image $\mathcal{I}$ so that the amplitude of the reference image can pay attention to specific regions in the input data. Cross-attention \cite{Chen-CVPR2021, Rombach-CVPR2022, le2025inftybrush} between an input image and a generated image can help to model the spatial correspondence of different source images. This can help to extract appropriate features and achieve adaptive and balanced fusion. It can dynamically learn the spatial correspondence to derive better alignment of essential details from the two images $\mathcal{I}$ and $\mathcal{I}_G$. 

The input to the Cross Attention module is an input image $\mathcal{I}$ and a generated image $\mathcal{I}_G$, both of size $H \times W \times C$, where $H\times W$ is the spatial dimension and $C$ is the number of channels (e.g., 3 for RGB images).
Inspired by Vision Transformer \cite{Dosovitskiy-ICLR2021}, first, we split both images into patches of size $P \times P \times C$, resulting in $(H/P) \times (W/P)$ patches for each image. Each patch is then flattened into a vector of size $P^2C$ and linearly projected into a token of size $D$ using a learnable projection matrix $W_e$ of size $P^2C \times D$. This can be expressed mathematically as:
\begin{equation}
    \mathcal{I}^t = I^p * W_e \textrm{ and } \mathcal{I}_G^t = \mathcal{I}_G^p * W_e, 
\end{equation}
where $\mathcal{I}^t$ and $\mathcal{I}_G^t$ are the token representations of the input and generated images; $I^p$ and $\mathcal{I}_G^p$ are the patch representations of the input and generated images, respectively.

Next, we apply cross-attention between the two sets of tokens to allow the reference image to pay attention to specific regions of the input image. This can be expressed mathematically using the scaled dot-product attention mechanism:
\begin{equation}
\begin{split}
A &= \mathrm{softmax}((\mathcal{I}^t * W_q) * (\mathcal{I}_G^t * W_k)^T / \sqrt{D}) \\
O &= A * (\mathcal{I}_G^t * W_v),    
\end{split}    
\end{equation}
where $W_q$, $W_k$, and $W_v$ are learnable projection matrices of size $D \times D$ for the query, key, and value representations, respectively, and $A$ is the attention matrix that represents the attention paid by the generated image to each patch of the input image. The output $O$ is then linearly projected via a learnable projection matrix $W_d$ of size $D \times P^2C$ and reshaped into an image of size $H \times W \times C$ in the RGB domain. To ensure that the output $O$ remains in the RGB domain with pixel values in the range $[0, 255]$, we apply a scaling and clipping operation to the output before reshaping it into an image. This can be done by first normalizing the output to have zero mean and unit variance, then scaling it by a factor of $255$, and finally clipping the values to the range $[0, 255]$. The final result of the cross-attention module is a reference image $\mathcal{I}_R$.

\subsection{Adaptive Hybrid Swapping of Amplitude}
Inspired by the hybrid image \cite{Oliva-SIGGRAPH2006}, which mixes the low-spatial frequencies of one image with the high spatial frequencies of another image, we propose an adaptive hybrid swapping the amplitude of the input image $\mathcal{I}$ and the reference image $\mathcal{I}_R$. We transform the input image $\mathcal{I}$ and the reference image $\mathcal{I}_R$ via FFT. That of the reference image adaptively replaces the low-frequency part of the amplitude of the input image. 
We denote $\mathbb{M}_\beta$ as a binary mask matrix, whose value is one in the center region and zero in the rest where $\beta \in (0, 1)$:
\begin{equation}
    \mathbb{M}_\beta(h, w) = \mathds{1}_{(h, w)\in [-\beta H:\beta H, -\beta W: \beta W]}.
\end{equation}

The inverse Fourier transform of the adaptive hybrid swapping of amplitude is given by:
\begin{equation}
\resizebox{\hsize}{!}{$
\mathcal{I}* = \mathcal{F}^{-1}\Big(\Big[\mathbb{M}_\beta \odot \mathcal{A}(\mathcal{F}(\mathcal{I}_R)) + (1 - \mathbb{M}_\beta) \odot \mathcal{A}(\mathcal{F}(\mathcal{I}))\Big], \phi(\mathcal{F}(\mathcal{I}))\Big).$}
\end{equation}

The parameter $\beta$ is optimized via Bayesian optimization. Consequently, the augmented image $\mathcal{I}*$ is the output of the adaptive hybrid swapping operation. The benefit of adaptive hybrid swapping is that it can control the amount of texture and color information that is transferred from the reference image $\mathcal{I}_R$ to the input image $\mathcal{I}$. The augmented image $\mathcal{I}*$ is fed as input to camouflaged object detection and instance segmentation models for training with specific objective functions of COD and CIS tasks.
\begin{table}[!t]
\centering
\resizebox{\linewidth}{!}{
\begin{tabular}{lcccccc}
\toprule
\multirow{2}{*}{\textbf{Method}} & \multicolumn{3}{c}{\textbf{COD10K-Test}} & \multicolumn{3}{c}{\textbf{NC4K-Test}} \\ 
 & 
  \textbf{AP} &
  \textbf{AP$_{50}$} &
  \textbf{AP$_{75}$} &
  \textbf{AP} & 
  \textbf{AP$_{50}$} &
  \textbf{AP$_{75}$} \\
\midrule
Mask R-CNN \cite{He-ICCV2017} & 0.250 & 0.555 & 0.204 & 0.277 & 0.586 & 0.227 \\
Mask R-CNN \cite{He-ICCV2017} + \textbf{\texttt{CamoFA}} & \blue{0.287} & \blue{0.591} & \blue{0.229} & \blue{0.305} & \blue{0.627} & \blue{0.251} \\
Cascade R-CNN \cite{Cai-CVPR2018} & 0.253 & 0.561 & 0.213 & 0.295 & 0.608 & 0.248 \\ 
Cascade R-CNN \cite{Cai-CVPR2018} + \textbf{\texttt{CamoFA}} & \blue{0.287} & \blue{0.589} & \blue{0.229} & \blue{0.313} & \blue{0.631} & \blue{0.264}\\ 
YOLACT \cite{Bolya-ICCV2019} & 0.243 & 0.533 & 0.197 & 32.10 & 0.653 & 0.279 \\ 
YOLACT \cite{Bolya-ICCV2019} + \textbf{\texttt{CamoFA}} & \blue{0.277} & \blue{0.574} & \blue{0.222} & \blue{0.345} & \blue{0.692} & \blue{0.308} \\ 
SOTR \cite{Guo-ICCV2021} & 0.279 & 0.587 & 0.241 & 0.293 & 0.610 & 0.256 \\
SOTR \cite{Guo-ICCV2021} + \textbf{\texttt{CamoFA}} & \blue{0.302} & \blue{0.616} & \blue{0.259} & \blue{0.326} & \blue{0.651} & \blue{0.289} \\
OSFormer \cite{Pei-ECCV2022} & 0.410 & 0.711 & 0.408 & 0.425 & 0.725 & 0.423 \\
OSFormer \cite{Pei-ECCV2022} + \textbf{\texttt{CamoFA}} & \blue{0.435} & \blue{0.748} & \blue{0.427} & \blue{0.450} & \blue{0.757} & \blue{0.443} \\

\bottomrule
\end{tabular}
}
\caption{Effectiveness of our \texttt{CamoFA} in improving CIS model OSFormer \cite{Pei-ECCV2022} and generic instance segmenters on two benchmarks: COD10K \cite{Fan-CVPR2020} and NC4K \cite{Lv-CVPR2021}. Our \texttt{CamoFA} significantly boosts the performance of state-of-the-art methods and improved results are highlighted in \blue{blue} color.} 
\label{tab:cis_exp}
\vspace{-2mm}
\end{table}
\section{Experiments}
\subsection{Experimental Settings}
\textbf{Datasets.}
For COD, we use 3 popular benchmarks: CAMO \cite{LE-CVIU2019}, COD10K \cite{Fan-CVPR2020}, and NC4K \cite{Lv-CVPR2021}. Among them, CAMO consists of 2,500 images, half with camouflaged objects and half without. COD10K \cite{Fan-CVPR2020} contains 5,066 camouflaged, 3,000 background, and 1,934 non-camouflaged images. NC4K is a large-scale COD dataset with 4,121 images. We follow prior studies  \cite{Fan-CVPR2020,Lv-CVPR2021} and use 1,000 images from CAMO \cite{LE-CVIU2019}, 3,040 images from COD10K\cite{Fan-CVPR2020}, and all images from NC4K \cite{Lv-CVPR2021} for testing.

For CIS, there are few task-specific datasets since this is a new and challenging task. Fan \etal \cite{Fan-CVPR2020} provided a COD dataset called COD10K that also has instance-level annotations for CIS models. COD10K has 2,026 images for testing and 3,040 camouflaged images with instance-level labels for training. Recently, Le \etal \cite{ltnghia-TIP2022} released the CAMO++ dataset, a larger CIS dataset with 5,500 samples with hierarchical pixel-wise annotation. Lyu \etal \cite{Lv-CVPR2021} introduced the NC4K test set for CIS, which has 4,121 images.

\begin{table*}[!t]
\centering
\footnotesize
\resizebox{1\linewidth}{!}{
\begin{tabular}{llcccccccccccc}
\toprule
\multirow{2}{*}{\textbf{Method}} & \multirow{2}{*}{\textbf{Augmentation}} & \multicolumn{4}{c}{\textbf{NC4K-Test}} & \multicolumn{4}{c}{\textbf{COD10K-Test}} &  \multicolumn{4}{c}{\textbf{CAMO-Test}} \\
 & & 
  S$_m\uparrow$ &
  $\alpha$E $ \uparrow$ &
  $w$F $ \uparrow$ &
  M $ \downarrow$ & 
  S$_m \uparrow$ &
  $\alpha$E $ \uparrow$ &
  $w$F $ \uparrow$ &
  M $ \downarrow$ &
  S$_m \uparrow$ &
  $\alpha$E $ \uparrow$ &
  $w$F $ \uparrow$ &
  M $ \downarrow$ \\
\midrule
\multirow{6}{*}{SINetV2 \cite{Fan-TPAMI2022}} & - & 0.847 & 0.898 & 0.770 & 0.048 & 0.815 & 0.863 & 0.680 & 0.037 & 0.820 & 0.875 & 0.743 & 0.070 \\ 
 & Mixup \cite{Zhang-ICLR2018} & 0.813 & 0.856 & 0.745 & 0.067 & 0.764 & 0.829 & 0.636 & 0.042 & 0.784 & 0.817 & 0.694 & 0.088 \\ 
 & Cutmix \cite{Yun-ICCV2019} & 0.806 & 0.838 & 0.723 & 0.071 & 0.752 & 0.816 & 0.630 & 0.056 & 0.771 & 0.805 & 0.683 & 0.079 \\ 
 & AutoAugment \cite{Zoph-ECCV2020} & 0.839 & 0.903 & 0.766 & 0.051 & 0.812 & 0.858 & 0.675 & 0.040 & 0.823 & 0.872 & 0.746 & 0.074 \\ 
 & Copy-Paste \cite{Ghiasi-CVPR2021} & 0.820 & 0.861 & 0.754 & 0.055 & 0.806 & 0.849 & 0.673 & 0.042 & 0.811 & 0.864 & 0.727 & 0.080 \\ 
 & CamDiff \cite{Luo-CamDiff2023} & 0.851 & 0.895 & 0.772 & 0.047 & 0.819 & 0.866 & 0.678 & 0.037 & 0.831 & 0.882 & 0.745 & 0.073 \\ 
& \textbf{\texttt{CamoFA}} & \textbf{0.859} & \textbf{0.914} & \textbf{0.778} & \textbf{0.043} & \textbf{0.826} & \textbf{0.890} & \textbf{0.704} & \textbf{0.032} & \textbf{0.858} & \textbf{0.893} & \textbf{0.752} & \textbf{0.066} \\ 
\bottomrule
\end{tabular}
}
\caption{Comparisons between our \texttt{CamoFA} and existing generic augmentation methods in the camouflaged object detection task. Cutting-edge augmentations show ineffectiveness when applying to COD while our \texttt{CamoFA} boosts the performance of SINetV2 \cite{Fan-TPAMI2022} significantly.}
\label{tab:cod_augment}
\vspace{-2mm}
\end{table*}

\begin{table}[!t]
\centering
\small
 \resizebox{\linewidth}{!}{
\begin{tabular}{llcccccc}
\toprule
\multirow{2}{*}{\textbf{Method}} & \multirow{2}{*}{\textbf{Augmentation}} & \multicolumn{3}{c}{\textbf{COD10K-Test}} & \multicolumn{3}{c}{\textbf{NC4K-Test}} \\ 
 & &
  \textbf{AP} &
  \textbf{AP$_{50}$} &
  \textbf{AP$_{75}$} &
  \textbf{AP} & 
  \textbf{AP$_{50}$} &
  \textbf{AP$_{75}$} \\
\midrule 
\multirow{6}{*}{OSFormer \cite{Pei-ECCV2022}} & - & 0.410 & 0.711 & 0.408 & 0.425 & 0.725 & 0.423 \\
 & Mixup \cite{Zhang-ICLR2018} & 0.397 & 0.685 & 0.386 & 0.392 & 0.681 & 0.387 \\
 & Cutmix \cite{Yun-ICCV2019} & 0.379 & 0.663 & 0.371 & 0.390 & 0.672 & 0.383 \\
 & AutoAugment \cite{Zoph-ECCV2020} & 0.413 & 0.715 & 0.393 & 0.427 & 0.723 & 0.426 \\ 
 & Copy-Paste \cite{Ghiasi-CVPR2021} & 0.398 & 0.713 & 0.386 & 0.419 & 0.716 & 0.414 \\
 & CamDiff \cite{Luo-CamDiff2023} & 0.422 & 0.726 & 0.414 & 0.433 & 0.729 & 0.422 \\ 
 & \textbf{\texttt{CamoFA}} & \textbf{0.435} & \textbf{0.748} & \textbf{0.427} & \textbf{0.450} & \textbf{0.757} & \textbf{0.443} \\
\bottomrule
\end{tabular}
 }
\caption{Comparisons between our \texttt{CamoFA} and existing generic augmentation methods in  CIS task. State-of-the-art generic augmentations downgrade the effectiveness of OSFormer \cite{Pei-ECCV2022} while our \texttt{CamoFA} drastically improves CIS models' performance.}
\label{tab:cis_augment}
\vspace{-2mm}
\end{table} 
\textbf{Evaluation metrics.}
For COD, we use four common metrics: Structure-measure (S$_m$) \cite{Fan-ICCV2017}, mean absolute error (M) \cite{Federico-CVPR2012}, weighted F-measure ($w$F) \cite{Margolin-CVPR2014}, and adaptive E-measure ($\alpha$E) \cite{Fan-IJCAI2018}. S$_m$ measures the structural similarity between predictions and ground truth, considering both region- and object-awareness. M measures the pixel-level accuracy between a predicted map and ground truth and is widely used in salient object detection (SOD) tasks. $w$F measures the balance between recall and precision. $\alpha$E measures the human visual perception-based alignment between predictions and ground truth at both pixel and image levels and is suitable for evaluating the global and local accuracy of camouflaged object detection results. Regarding CIS, we adopt COCO-style \cite{Lin-ECCV2014} evaluation metrics such as AP$_{50}$, AP$_{75}$, and AP. Unlike the mAP metric used in instance segmentation, we do not consider the class labels of camouflaged instances, since they are class-agnostic. We only need to consider the existence of camouflaged instances and ignore the class mean value.

\subsection{Implementation Details}
In our \texttt{CamoFA}, the GAN architecture consists of generator and discriminator components. We adopt a simple UNet encoder-decoder architecture for the generator.  
However, we remove all skip connections between the encoder and the decoder. Regarding the discriminator, we leverage $70 \times 70$ PatchGAN \cite{Isola-CVPR2017} to classify if each $N \times N$ patch in an image is real or fake. 
The Fast Fourier Transform (FFT) and iFFF are implemented using the PyTorch FFT package. Our \texttt{CamoFA} is plugged into several camouflaged object detectors and instance segmenters for end-to-end training. The parameter $\beta$ in adaptive hybrid swapping is optimized by Bayesian optimization in range $(0, 1)$ with step size $0.01$. Original images are resized to $512 \times 512$ and transformed using \texttt{CamoFA} then fed into specific architectures of COD and CIS. We adopt the Adam optimizer with a learning rate of $10^{-4}$, $\beta_1 = 0.5$, and $\beta_2 = 0.999$. All variants of \texttt{CamoFA} are trained with a batch size of $8$ on a single NVIDIA RTX 3090 GPU.

\subsection{Comparisons with State-of-the-art Methods}
We validate the effectiveness of our learnable augmentation in the frequency domain by integrating our augmentation strategy into several camouflaged object detection and camouflaged instance segmentation models. 

\textbf{Camouflaged object detection.} We integrate the proposed \texttt{CamoFA} into $5$ different state-of-the-art camouflaged object detection models including: SINet \cite{Fan-CVPR2020}, SINetV2 \cite{Fan-TPAMI2022}, SegMaR \cite{Jia-CVPR2022}, ZoomNet \cite{Pang-CVPR2022}, and FDNet \cite{Zhong-CVPR2022}. The experimental results presented in Table~\ref{tab:cod_exp} show \texttt{CamoFA} improves the performance of the 5 methods on all considered metrics.

\textbf{Camouflaged instance segmentation.} The CIS task has not been well-studied as only one CIS model (OSFormer \cite{Pei-ECCV2022}) has been proposed to date. Therefore, we also conduct an experiment on several popular generic instance segmentation models for more comprehensive evaluation. Following the experiments of OSFormer \cite{Pei-ECCV2022}, we train those instance segmentation models on the instance-level COD10K training set and test them on the COD10K \cite{Fan-CVPR2020} and NC4K \cite{Lv-CVPR2021} test sets. Table \ref{tab:cis_exp} shows that our learnable augmentation improves the performance of the instance segmentation models in the CIS task on all AP metrics. Specifically, our method increases the AP score of OSFormer \cite{Pei-ECCV2022} by a significant margin of $2.52$. For the generic instance segmentation model Mask R-CNN \cite{He-ICCV2017}, our augmentation method boosts the AP score by an even higher margin of \textbf{$3.74$}.

\textbf{Comparisons with state-of-the-art augmentation methods.} We compare our \texttt{CamoFA} with CamDiff \cite{Luo-CamDiff2023} and cutting-edge augmentation methods for generic object detection and instance segmentation including Mixup \cite{Zhang-ICLR2018}, CutMix \cite{Yun-ICCV2019}, AutoAugment \cite{Zoph-ECCV2020}, and Copy-Paste \cite{Ghiasi-CVPR2021} on COD and CIS tasks. In particular, we plug the mentioned augmentation methods into SINetV2 \cite{Fan-TPAMI2022} and OSFormer \cite{Pei-ECCV2022} for COD and CIS tasks, respectively. Table~\ref{tab:cod_augment} and Table~\ref{tab:cis_augment} indicate that generic augmentation methods are obviously not suitable for camouflaged object detection and instance segmentation; they even make the performance of SINetV2 \cite{Fan-TPAMI2022} and OSFormer \cite{Pei-ECCV2022} deteriorate compared to the ones without augmentation. On the contrary, our proposed \texttt{CamoFA} helps improve the effectiveness of SINetV2 \cite{Fan-TPAMI2022} and OSFormer \cite{Pei-ECCV2022} by large margins. \texttt{CamoFA} also outperforms CamDiff \cite{Luo-CamDiff2023}, the current state-of-the-art method for camouflaged objects on COD and CIS tasks.
\subsection{Qualitative Results}

\begin{figure}[t!]
    \centering
    \includegraphics[width=\linewidth]{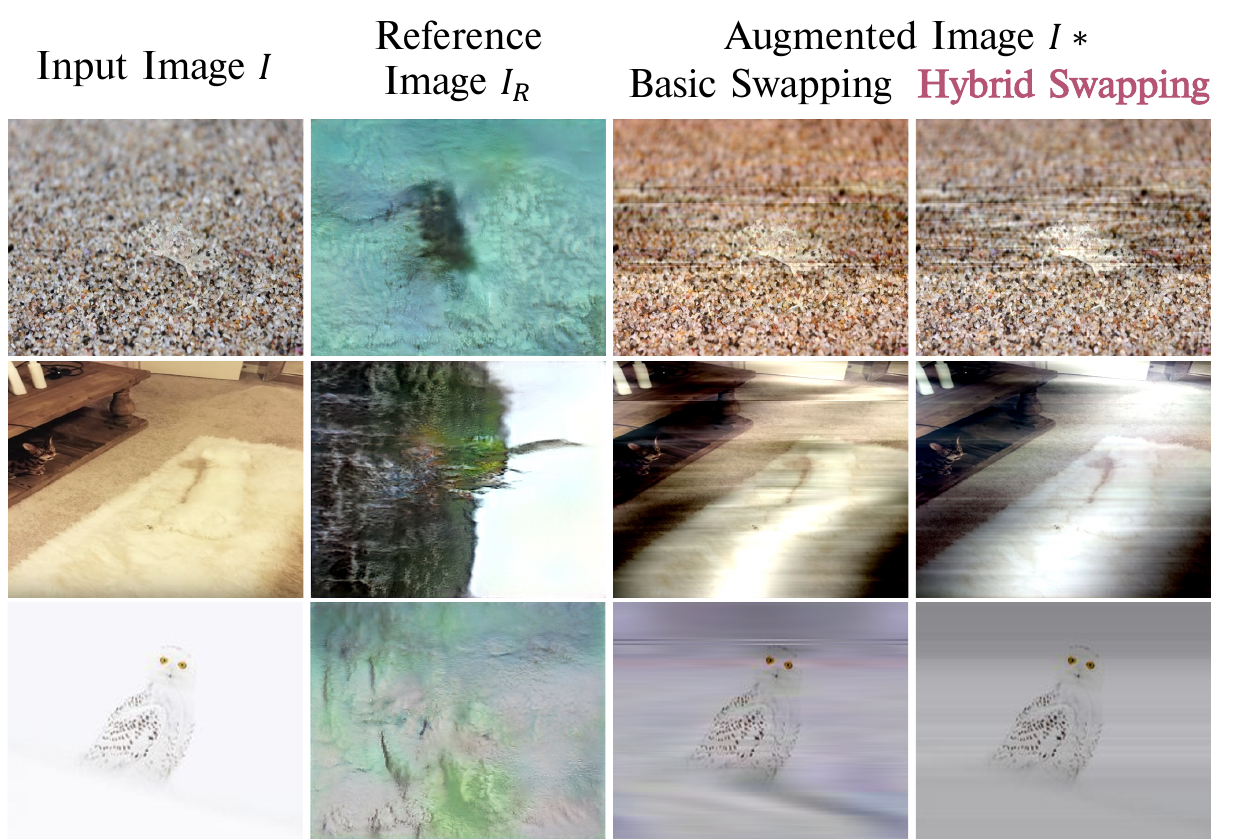}
    \caption{Visualization of augmented images by our \texttt{CamoFA}. Our proposed augmentation highlights the underlying structure of camouflaged objects for better identification and segmentation.  We also compare the transformed results of our method with and without adaptive hybrid swapping. Our adaptive hybrid swapping can control the amount of texture and color information that is transferred from the reference image to the input image.}
    \label{fig:augmented_img}
    \vspace{-2mm}
\end{figure}

\begin{figure}[!t]
    \centering
    \includegraphics[width=\linewidth]{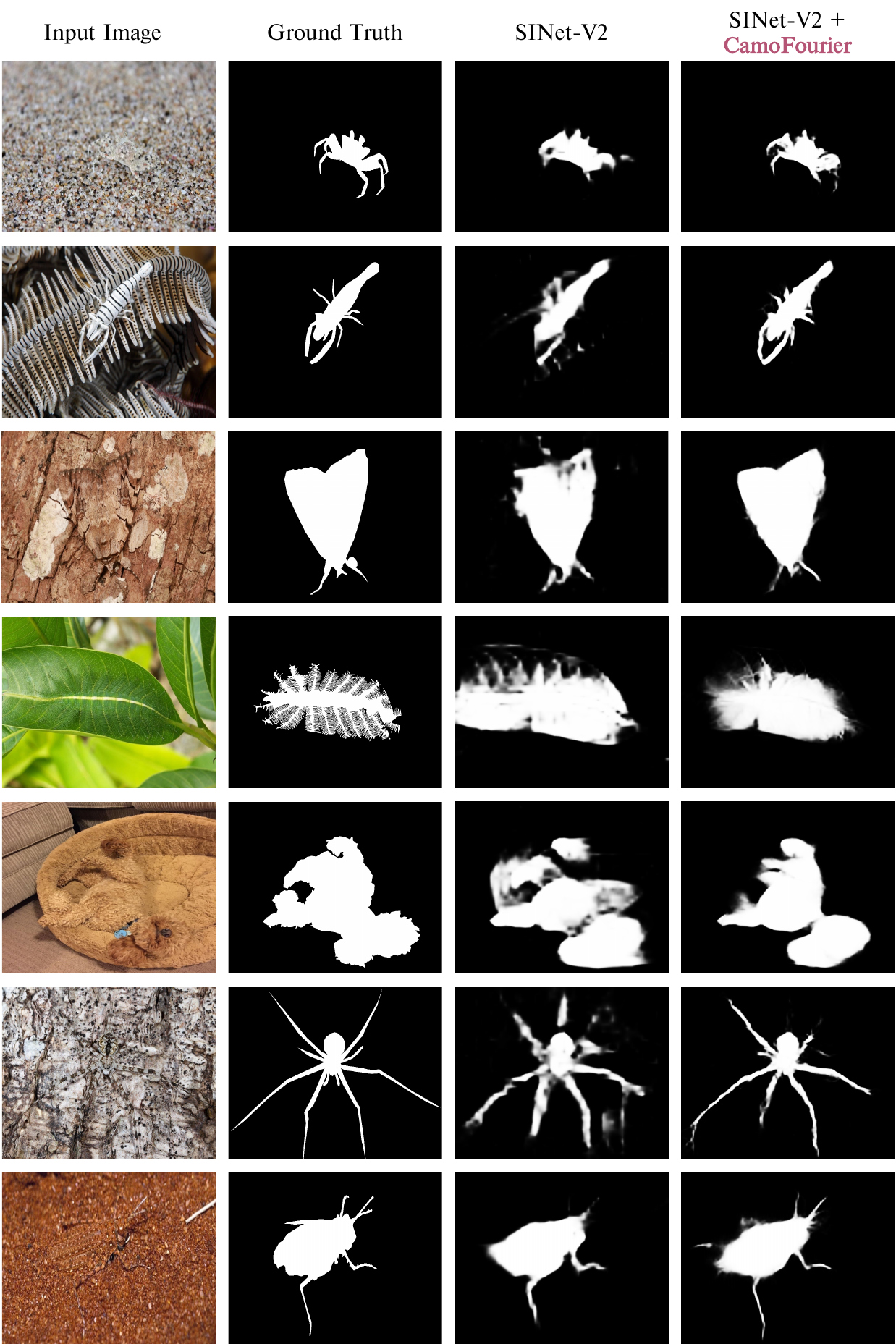}
    \caption{Qualitative comparison of SINetV2 with and without our proposed \texttt{CamoFA} in COD task.}
    \label{fig:qualitative_cod}
    \vspace{-2mm}
\end{figure}

\textbf{Visualization of augmented images.} In Figure~\ref{fig:augmented_img}, we illustrate the outputs of our \texttt{CamoFA}, which are augmented images $\mathcal{I}*$. In the \texttt{CamoFA} framework, the conditional GAN module and a cross-attention mechanism synthesize a reference image $\mathcal{I}_R$. After that, we transform an input image and a reference image into the frequency domain and perform basic swapping or adaptive hybrid swapping of these two amplitudes. We also compare the qualitative results of our \texttt{CamoFA} with and without adaptive hybrid swapping. Figure~\ref{fig:augmented_img} shows that our adaptive hybrid swapping can control the amount of texture and color information that is transferred from the reference image $\mathcal{I}_R$ to the input image $\mathcal{I}$ thanks to the parameter $\beta$.

\begin{figure}[!t]
    \centering
    \includegraphics[width=\linewidth]{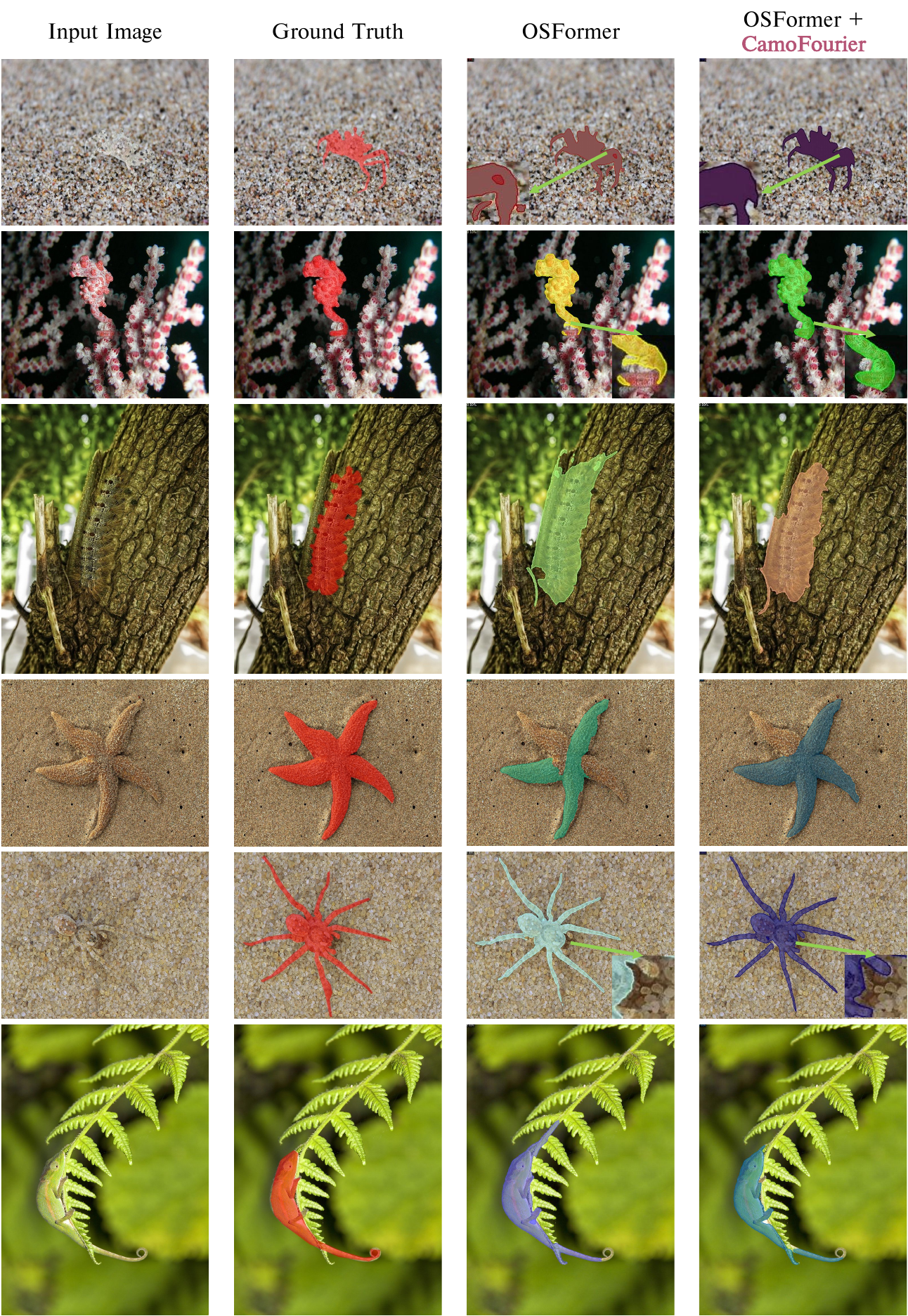}
    \caption{Qualitative comparison of OSFormer with and without our proposed \texttt{CamoFA} in CIS task.}
    \label{fig:qualitative_cis}
    \vspace{-3mm}
\end{figure}

\textbf{Qualitative evaluation.}
Figures~\ref{fig:qualitative_cod} and~\ref{fig:qualitative_cis} visualize segmentation results of SINetV2 \cite{Fan-TPAMI2022} and OSFormer \cite{Pei-ECCV2022} with and without our \texttt{CamoFA}, respectively. The results demonstrate SINetV2 and OSFormer equipped with \texttt{CamoFA} can delineate camouflaged objects more accurately. 
\begin{table*}[!t]
\centering
\footnotesize
\resizebox{\linewidth}{!}{
\begin{tabular}{lcccccccccc}
\toprule
\multirow{2}{*}{\textbf{Method}} & \multicolumn{2}{c}{\textbf{\texttt{CamoFA}}} & \multicolumn{4}{c}{\textbf{NC4K-Test}} & \multicolumn{4}{c}{\textbf{COD10K-Test}} \\ 
 & Cross-attention & Hybrid swapping &
  S$_m\uparrow$ &
  $\alpha$E $ \uparrow$ &
  $w$F $ \uparrow$ &
  M $ \downarrow$ & 
  S$_m \uparrow$ &
  $\alpha$E $ \uparrow$ &
  $w$F $ \uparrow$ &
  M $ \downarrow$ \\
\midrule
\multirow{4}{*}{SINetV2 \cite{Fan-TPAMI2022}} & \ding{55} & \ding{55} & 0.852 & 0.908 & 0.773 & 0.045 & 0.821 & 0.882 & 0.695 & 0.034 \\
& \ding{51} & \ding{55} & 0.857 & 0.911 & 0.776 & 0.044 & 0.824 & 0.888 & 0.697 & 0.033 \\
& \ding{55} & \ding{51} & 0.853 & 0.910 & 0.774 & 0.045 & 0.825 & 0.884 & 0.699 & 0.033 \\
& \ding{51} & \ding{51} & \textbf{0.859} & \textbf{0.914} & \textbf{0.778} & \textbf{0.043} & \textbf{0.826} & \textbf{0.890} & \textbf{0.704} & \textbf{0.032} \\
\midrule
% 0.853 & 0.907 & 0.784 & 0.043 & 0.838 & 0.893 & 0.729 & 0.029
\multirow{4}{*}{ZoomNet \cite{Pang-CVPR2022}} & \ding{55} & \ding{55} & 0.868 & 0.915 & 0.792 & 0.041 & 0.851 & 0.907 & 0.735 & 0.027 \\
& \ding{51} & \ding{55} & 0.871 & 0.920 & 0.798 & 0.039 & 0.860 & 0.910 & 0.738 & 0.026 \\
& \ding{55} & \ding{51} & 0.869 & 0.917 & 0.793 & 0.037 & 0.855 & 0.910 & 0.737 & 0.027 \\
& \ding{51} & \ding{51} & \textbf{0.872} & \textbf{0.923} & \textbf{0.801} & \textbf{0.037} & \textbf{0.864} & \textbf{0.911} & \textbf{0.740} & \textbf{0.025} \\
\bottomrule
\end{tabular}
}
\caption{Our ablation study in the task of camouflaged object detection. \texttt{CamoFA} with a cross-attention mechanism and adaptive hybrid swapping achieves the best performance when plugged into SINetV2 \cite{Fan-TPAMI2022} and ZoomNet \cite{Pang-CVPR2022}.}
\vspace{-3mm}
\label{tab:cod_ablation}

\end{table*}
\begin{table}[!t]
\centering
\footnotesize
\resizebox{\linewidth}{!}{
\begin{tabular}{lcccccccc}
\toprule
\multirow{2}{*}{\textbf{Method}} & \multicolumn{2}{c}{\textbf{\texttt{CamoFA}}} & \multicolumn{3}{c}{\textbf{COD10K-Test}} & \multicolumn{3}{c}{\textbf{NC4K-Test}} \\ 
 & Cross-attention & Hybrid swapping &
  \textbf{AP} &
  \textbf{AP$_{50}$} &
  \textbf{AP$_{75}$} &
  \textbf{AP} & 
  \textbf{AP$_{50}$} &
  \textbf{AP$_{75}$} \\
\midrule
\multirow{4}{*}{Mask R-CNN \cite{He-ICCV2017}} & \ding{55} & \ding{55} & 0.276 & 0.578 & 0.216 & 0.292 & 0.616 & 0.248 \\
& \ding{51} & \ding{55} & 0.283 & 0.586 & 0.222 & 0.301 & 0.624 & 0.250 \\
& \ding{55} & \ding{51} & 0.278 & 0.583 & 0.219 & 0.297 & 0.621 & 0.249 \\
& \ding{51} & \ding{51} & \textbf{0.288} & \textbf{0.591} & \textbf{0.229} & \textbf{0.305} & \textbf{0.627} & \textbf{0.251} \\
\midrule
\multirow{4}{*}{OSFormer \cite{Pei-ECCV2022}} & \ding{55} & \ding{55} & 0.424 & 0.729 & 0.420 & 0.439 & 0.737 & 0.430\\
& \ding{51} & \ding{55} & 0.430 & 0.742 & 0.424 & 0.446 & 0.752 & 0.440 \\
& \ding{55} & \ding{51} & 0.428 & 0.738 & 0.423 & 0.441 & 0.743 & 0.437 \\
& \ding{51} & \ding{51} & \textbf{0.435} & \textbf{0.749} & \textbf{0.427} & \textbf{0.450} & \textbf{0.757} & \textbf{0.443} \\
\bottomrule
\end{tabular}
}
\caption{Our ablation study in the task of camouflaged instance segmentation. \texttt{CamoFA} with cross-attention mechanism and adaptive hybrid swapping achieves the best performance when plugged into OSFormer \cite{Pei-ECCV2022} and Mask R-CNN \cite{He-ICCV2017}.}
\label{tab:cis_ablation}
\vspace{-3mm}
\end{table}

\subsection{Ablation Study}
\textbf{Effectiveness of cross-attention module.} We evaluate the effectiveness of our cross-attention module within the \texttt{CamoFA} framework by comparing the performance of COD models (SINetV2 \cite{Fan-TPAMI2022}, ZoomNet \cite{Pang-CVPR2022}) and CIS models (OSFormer \cite{Pei-ECCV2022}, Mask R-CNN \cite{He-ICCV2017}) with and without the cross-attention mechanism. Without cross-attention, we directly transform the input image $\mathcal{I}$ and the GAN-generated image $\mathcal{I}_G$ into the frequency domain for amplitude swapping. As shown in Table~\ref{tab:cod_ablation}, the cross-attention module improves COD performance on the COD10K \cite{Fan-CVPR2020} and NC4K \cite{Lv-CVPR2021} test sets across four metrics. Similarly, Table~\ref{tab:cis_ablation} demonstrates that cross-attention enhances CIS performance on these test sets.

\textbf{Significance of adaptive hybrid swapping.} We evaluate the significance of the adaptive hybrid swapping in \texttt{CamoFA}. We also perform experiments on COD and CIS tasks with two models: SINetV2 \cite{Fan-TPAMI2022} and ZoomNet \cite{Pang-CVPR2022} for COD, 
OSFormer \cite{Pei-ECCV2022} and Mask R-CNN \cite{He-ICCV2017} for CIS. 
We apply our \texttt{CamoFA} to these architectures with and without adaptive hybrid swapping. Without using the proposed hybrid swapping, we use the basic swapping of amplitude as described in Section~\ref{paragraph:basic_swap}. Tables~\ref{tab:cod_ablation} and~\ref{tab:cis_ablation} show that our adaptive hybrid swapping helps the \texttt{CamoFA} augmentation to boost the performance of COD and CIS models.

\section{Conclusion}
In this paper, we propose \texttt{CamoFA}, a novel learnable Fourier-based augmentation method for camouflaged object detection and instance segmentation. Our method incorporates a generative model and cross-attention mechanism to create a reference image and a Fourier transform to mix the low and high-frequency components of the input and reference images. This makes the camouflaged objects more visible and easier to detect and segment. %Our method improves the performance of existing models for both tasks significantly. 
We conduct extensive experiments on various camouflaged object detection and instance segmentation models on different datasets. The experimental results show that our method significantly improves the existing state-of-the-art COD and CIS methods by a large margin. 

\paragraph{Acknowledgments.} This research is partially supported by Vingroup Innovation Foundation (VINIF) in project code VINIF.2019.DA19 and National Science Foundation (NSF) under Grant No. 2025234.
%%%%%%%%% REFERENCES
{\small
\bibliographystyle{ieee_fullname}
\bibliography{egbib}
}

\end{document}